\documentclass[11pt]{article}
\usepackage{coling2020}
\usepackage{tcolorbox}
\usepackage{latexsym}
\usepackage{hyperref}
\usepackage{multirow}
\usepackage{graphicx}
\usepackage{amsmath}
\usepackage{subfig}
\usepackage{times}
\usepackage{url}

\colingfinalcopy

\title{ECSP: A New Task for Emotion-Cause Span-Pair \\Extraction and Classification}

\author{Hongliang Bi, Pengyuan Liu \\
  Beijing Language and Culture University, China\\
  {\tt 201821198617@stu.blcu.edu.cn, liupengyuan@blcu.edu.cn} \\}

\date{}

\begin{document}
\maketitle
\begin{abstract}
Emotion cause analysis such as emotion cause extraction (ECE) and emotion-cause pair extraction (ECPE) have gradually attracted the attention of many researchers. However, there are still two shortcomings in the existing research: 1) In most cases, emotion expression and cause are not the whole clause, but the span in the clause, so extracting the clause-pair rather than the span-pair greatly limits its applications in real-world scenarios; 2) It is not enough to extract the emotion expression clause without identifying the emotion categories, the presence of emotion clause does not necessarily convey emotional information explicitly due to different possible causes. 
In this paper, we propose a new task: Emotion-Cause Span-Pair extraction and classification (ECSP), which aims to extract the potential span-pair of emotion and corresponding causes in a document, and make emotion classification for each pair. In the new ECSP task, ECE and ECPE can be regarded as two special cases at the clause-level. We propose a span-based extract-then-classify (ETC) model, where emotion and cause are directly extracted and paired from the document under the supervision of target span boundaries, and corresponding categories are then classified using their pair representations and localized context. 
Experiments show that our proposed ETC model outperforms the SOTA model of ECE and ECPE task respectively and gets a fair-enough results on ECSP task.
\end{abstract}

\section{Introduction}
\begin{NoHyper}
\blfootnote{
    %
    % for review submission
    %
    \hspace{-0.65cm}  % space normally used by the marker
    Preprint version.
    % 
    % % final paper: en-us version 
    %
    % \hspace{-0.65cm}  % space normally used by the marker
    % This work is licensed under a Creative Commons 
    % Attribution 4.0 International License.
    % License details:
    % \url{http://creativecommons.org/licenses/by/4.0/}.
}
\end{NoHyper}
Emotion cause analysis such as emotion cause extraction (ECE) and emotion-cause pair extraction (ECPE) have gradually attracted the attention of many researchers, can be constructive to guide the direction of future work, i.e., improving the quality of products or services according to the emotion causes of comments provided by users.

Emotion cause extraction (ECE) was first proposed by \newcite{lee2010text}, which aims at discovering the potential cause clauses behind a certain emotion expression in the text. Earlier work viewed ECE as a trigger word detection problem and tries to solve it with corresponding tagging techniques. Therefore, primary efforts have been made on discovering refined linguistic features \cite{chen2010emotion,lee2013detecting}, yielding improved performance. More recently, instead of concentrating on word-level cause detection, clause-level extraction \cite{gui2016event} was putted forward in that the impact of individual words in a clause can span over the whole sequence in the clause. While ECE has attracted an increasing attention due to its theoretical and practical significance, it requires that the emotion expression annotations should be given in the test set. In light of recent advances in multi-task learning, \newcite{chen2018joint} proposed joint extraction of emotion categories and causes are investigated to exploit the mutual information between two correlated tasks, and \newcite{xia2019emotion} proposed emotion-cause pair extraction (ECPE) task, which aims to extract all potential clause-pairs of emotion expression and corresponding cause in a document, and to solve the shortcomings of previous ECE task must be annotated before extraction causes. \newcite{xia2019emotion} argues that, while co-extraction of emotion expression and causes are important, ECPE is a more challenging problem that is worth putting more emphases on.

However, ECPE still suffers from two shortcomings: 1) In most cases, emotion expression and cause are not the whole clause, but the span in the clause, so extracting the clause-pair rather than the span-pair greatly limits its applications in real-world scenarios; 2) It is not enough to extract the emotion expression clause without identifying the emotion categories, the presence of emotion clause does not necessarily convey emotional information due to different possible causes such as negative polarity, sense ambiguity or rhetoric. For example, ``It feels like the sky is falling right on top of me'' is an emotion expression of ``fear''.

\begin{figure}[!tbp]
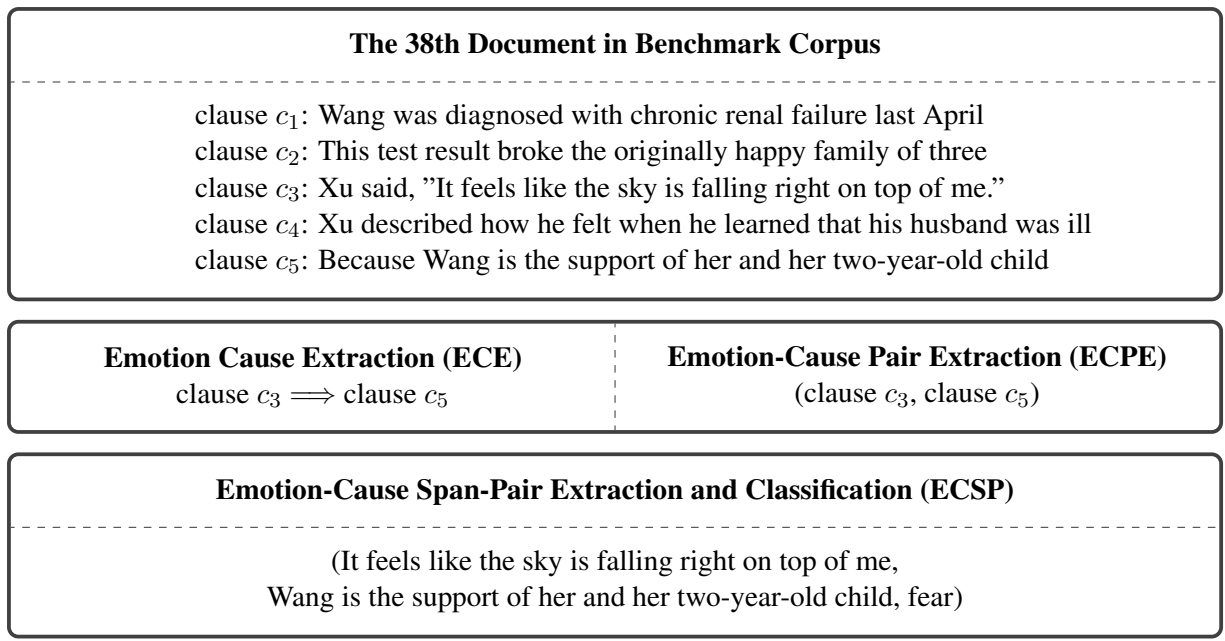

\centering
\begin{tcolorbox}[colback=white]
\centering
\textbf{The 38th Document in Benchmark Corpus}
\tcblower
\hspace{5em}clause $c_{1}$: Wang was diagnosed with chronic renal failure last April\par
\hspace{5em}clause $c_{2}$: This test result broke the originally happy family of three\par
\hspace{5em}clause $c_{3}$: Xu said, "It feels like the sky is falling right on top of me."\par
\hspace{5em}clause $c_{4}$: Xu described how he felt when he learned that his husband was ill\par
\hspace{5em}clause $c_{5}$: Because Wang is the support of her and her two-year-old child
\end{tcolorbox}
\begin{tcolorbox}[colback=white, sidebyside]
\centering
\textbf{Emotion Cause Extraction (ECE)}\\
clause $c_{3}$ $\Longrightarrow$ clause $c_{5}$
\tcblower
\centering
\textbf{Emotion-Cause Pair Extraction (ECPE)}\\
(clause $c_{3}$, clause $c_{5}$)
\end{tcolorbox}
\begin{tcolorbox}[colback=white]
\centering
\textbf{Emotion-Cause Span-Pair Extraction and Classification (ECSP)}
\tcblower
\centering
(It feels like the sky is falling right on top of me, \\Wang is the support of her and her two-year-old child, fear)
\end{tcolorbox}
\caption{An intuitive example of the difference between the ECE, ECPE and new ECSP task.}
\label{example} 
\end{figure}

In this paper, we propose a new task: Emotion-Cause Span-Pair extraction and classification (ECSP), which aims to extract the potential span-pair of emotion and corresponding causes in a document, and make emotion classification for each pair. Therefore, ECE and ECPE can be regarded as two special cases of ECSP at the clause-level. Figure \ref{example} is an intuitive example of the difference between the ECE, ECPE and new ECSP task.

Inspired by recent span-based models in syntactic parsing and co-reference resolution \cite{lee2017end,stern2017minimal}, we propose a span-based model to solve this new ECSP task. The key insight is to annotate each emotion and cause with its span boundary followed by its emotion categories. Under such annotation, we introduce a span-based extract-then-classify (ETC) model that emotion and cause are directly extracted and paired from the document under the supervision of target span boundaries, and corresponding categories are then classified using their pair representations and localized context. The advantage of this method is that clause-based tasks and span-based tasks can be interpreted uniformly. Moreover, since the polarity is decided by using the targeted span representation, the model is able to take all target words into account before making predictions, thus naturally avoiding sentiment inconsistency. 

We take BERT \cite{devlin2019bert} as the default backbone network, and explore the following two aspects. First, we explore the feasibility of the ECSP task under different length search schemes, and the results prove that the ECSP task can be solved well with the increase of the model search length, and there is still some room for improvement. Second, following previous works \cite{gui2016event,xia2019emotion}, we compare our proposed ETC model and strong baselines under the clause-based search scheme. our proposed ETC model outperforms the SOTA model of ECE and ECPE task respectively and gets a fair-enough results on ECSP task. This proves the feasibility of the ECSP task and the effectiveness of our proposed ETC model.

% The contributions of this paper are listed as follows:
% \begin{itemize}
% \item 
% \item 
% \item 
% \end{itemize}

\section{Proposed model}
\begin{figure}[!tbp]
\centering
\includegraphics[width=\textwidth]{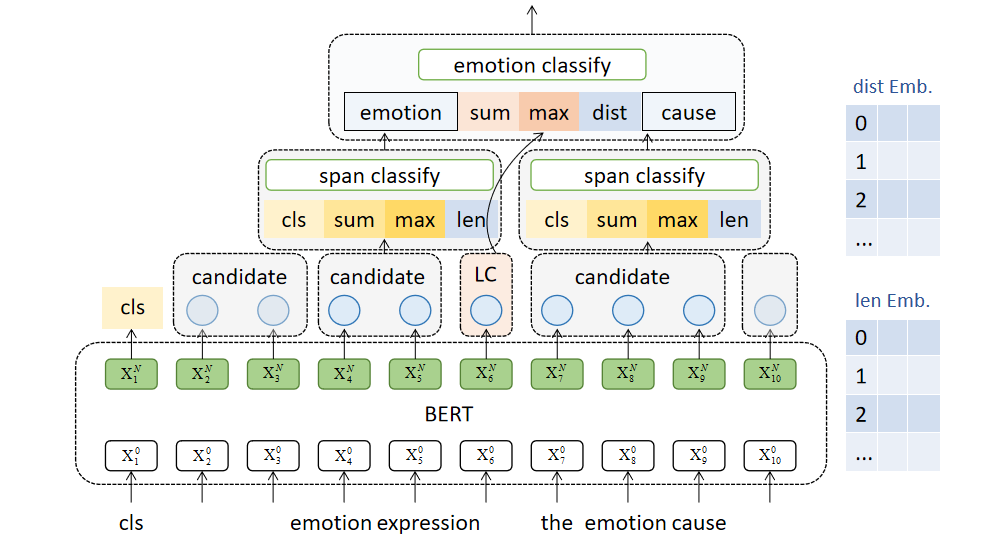}
\caption{Overall illustration of our proposed ETC model.}
\label{model}
\end{figure}
Instead of traditional clause-based detection methods to identify emotions and causes, we propose use a span-based search scheme as follows: give an input document $D = (x_{1}, \cdots, x_{n})$ with length $n$, and a emotion-cause span-pair list $P = \{p_{1}, \cdots, p_{m}\}$, where the number of emotion-cause span-pair is $m$ and each emotion expression span $e$ and corresponding cause span $c$ in pair $p_{i}$ is annotated with its START position, its END position, and its emotion category. Span $i$ is defined by all the tokens from START($i$) to END($i$) inclusive, for 1 $\leq$ $i$ $\leq$ N.

Our goal is to find all potential span-pair of emotion and corresponding causes in a document, and make emotion classification for each pair. The overall illustration of the proposed ETC model is shown in Figure \ref{model}. The basis of our proposed ETC model is the BERT encoder \cite{devlin2019bert}, we map word embeddings into contextualized token representations using pre-trained Transformer blocks \cite{vaswani2017attention}. A span classifier is first used to propose multiple candidate targets from the sentence. Then, an emotion classifier is designed to predict the emotion labels towards each extracted candidate span-pair using its summarized span representation and and localized context. We further study the performance of different span search schemes.

\subsection{Span Representation}
As mentioned before, we first obtain the features of tokens with BERT, which utilizes the abundant language knowledge, position information, and contextual information it contains. Given a document $D={\{x_{t}\}}$ where $t$ is the number of words, BERT begins by converting then sequence of tokens into a sequence of vectors $\mathbf{X}^{0}={\{x_{t}^{0}\}}^{L}_{i}$, $x^{0}_{t} \in \mathcal{R}^{d}$. Each of these vectors is the sum of a token embedding, a positional embedding that represents the position of the token in the sequence, and a segment embedding that represents whether the token is in the source text or the auxiliary text. We only have source text so the segment embeddings are the same for all tokens. Then several Transformer \cite{vaswani2017attention} layers are applied to get the final representations:
\begin{equation*}
\mathbf{X}^{i+1} = Transformer(\mathbf{X}^{i}), i \in [0, D-1]
\tag{1}
\end{equation*}
We use the final hidden output of BERT $\mathbf{X}^{D} \in \mathcal{R}^{L\times d}$ as the representations of corresponding tokens.

Attention mechanism \cite{bahdanau2015neural} can quickly extract important features of sparse data, so it is widely used in natural language. However, the BERT encoder uses a lot of attention mechanism, in order to save resources, it is no longer used. We use the following two convenient functions to create task-specific span features: (1) \textbf{sum} of all vectors for the entire span can usually represent the its semantics. (2) \textbf{max pooling} is a sample-based discretization process, which the objective is to down-sample an input representation (image, text, hidden-layer output matrix, etc.). For each span $i$, its span representation $\mathbf{g}_{i}$ was defined as:
\begin{equation*}
\mathbf{g}_{i} = concat(x^{D}_{cls}, sum(i), max(i), \Phi_{i})
\tag{2}
\end{equation*}
where $x^{D}_{cls}$ represents the final hidden output of BERT global context information, which is usually represented by the vector of the first token in BERT. $\Phi_{i}$ encodes the length of span $i$ in number of tokens. Each component of $\mathbf{g}_{i}$ is a span-specific feature that would be difficult to define and use in token-level models.

\subsection{Jointly Extract Emotion and Cause}
After obtaining span representation, we predict the type for each span. This prediction is done identically and parallelly for each span. For each span we compute a vector of type scores and apply the softmax function to its type score vector to obtain the distribution. For span $i$,
\begin{equation*}
y_{i}^{span} = softmax(\mathbf{g}_{i}w_{i} + b_{i})
\tag{3}
\label{eq3}
\end{equation*}
where $w_{i}$ and $b_{i}$ are parameters that can be learned.

The predicted type for each span $i$ is the type corresponding to span $i$'s highest span type score. Only spans whose predicted type is not none are selected. 

\subsection{Emotion-Cause Classification}
Finally, we obtain a set of emotion expression spans $E = \{\cdots, e_{i}, \cdots\}$ and a set of cause spans $C = \{\cdots,c_{i}, \cdots\}$. Now our goal is then to pair the two sets and construct a set of emotion-cause span-pairs with emotion relationship. Firstly, we apply a Cartesian product to $E$ and $C$, and obtain the set of all possible span-pairs:
\begin{equation*}
P_{all} = \{\cdots, (e_{i}, c_{j}), \cdots\}
\tag{4}
\end{equation*}

Despite advances in detecting long distance relations using BERT or the attention mechanism, the noise induced with increasing context remains a challenge. By using a Localized Context (LC), i.e. the context between span candidates, the emotion classifier can focus on the sentence’s section that is often most discriminative for the emotion type:
\begin{equation*}
\begin{aligned}
&p_{(e_{i}, c_{j})} = concat(\mathbf{g}_{e_{i}}, \mathbf{g}_{c_{j}}, LC_{(e_{i}, c_{j})})\\
LC_{(e_{i}, c_{j})} &= concat(sum(i\to j), max(i\to j), \Psi_{i\to j})
\end{aligned}
\tag{5}
\label{eq5}
\end{equation*}
where $e_{i}$ and $c_{j}$ are the representations of the emotion expression span and corresponding cause span respectively, $i\to j$ is localized context between $i$ and $j$, and $\Psi_{i\to j}$ represents the distances (dist) between span $i$ and span $j$.

For each emotion-cause span-pair $(e_{i}, c_{j})$, we obtain a representation by concatenating the respective span embeddings and Localized Context features. Finally, we train a softmax classifier to identify emotion categories:
\begin{equation*}
y_{(e_{i}, c_{j})}^{pair} = softmax(p_{(e_{i}, c_{j})}w_{(e_{i}, c_{j})} + b_{(e_{i}, c_{j})})
\tag{6}
\label{eq6}
\end{equation*}
where $w_{(e_{i}, c_{j})}$ and $b_{(e_{i}, c_{j})}$ are parameters that can be learned.

\subsection{Loss Function}
Two learning signals are provided to train this model: the span type information for each span (emotional expression, reason, and none) and the emotion category information for each selected (ordered) span-pair. Both are provided via cross-entropy \cite{shore1980axiomatic} loss on Eq. \ref{eq3} and Eq. \ref{eq6} respectively.

\section{Experiments}
\subsection{Corpus}
We evaluate on the benchmark ECE corpus\footnote{Available at: \url{http://www.hitsz-hlt.com/?page_id=694}} \cite{gui2016event}, which was the mostly used corpus for emotion cause extraction. The corpus includes annotations of emotional expressions and corresponding emotional causes. We use the boundary of the annotations as the start and end of the spans. Note that the presence of emotion expression does not necessarily convey emotional information due to different possible causes such as negative polarity, sense ambiguity or rhetoric. And, the presence of emotion expression does not necessarily guarantee the existence of emotional cause neither. Therefore, for each emotion expression, we also use the emotion labels provided by the corpus. There are different lengths for each emotion expression and cause, and the number is shown in Table \ref{corpus}.
\begin{table}[!tbp]
\centering
\begin{tabular}{lcccccc}
\hline
Item& Instance & Clauses & Cause & Cause\_1 & Cause\_2 & Cause\_3 \\ \hline
Number&  2105&  11799&  2167&  2046&  56& 3 \\ \hline
Item & Annotations & Length $\leq$ 2 & Length $\leq$ 5 & Length $\leq$ 10 & Length $\leq$ 15 & Length $\leq$ 20 \\ \hline
Number& 3879 & 1841 & 2587 & 3193 & 3655 & 3812 \\ \hline
\end{tabular}
\caption{Details of the corpus. Cause\_1, Cause\_2 and Cause\_3 represent the documents with 1, 2 and 3 causes, respectively. Length represent the length of each annotation.}
\label{corpus}
\end{table}

\subsection{Metrics}
The precision (P), recall (R), and F1 score are used as the metrics for evaluation. These metrics in emotion cause extraction are defined by:
\begin{equation*}
Precision = \frac{\sum correct\_items}{\sum proposed\_items},
Recall = \frac{\sum correct\_items}{\sum annotated\_items},
F1 = \frac{2 * Precision * Recall}{Precision + Recall}
\end{equation*}
where $proposed\_items$ denotes the number of items that are predicted, $annotated\_items$ denotes the number of items that in corpus and the $correct\_items$ means the number of items that are correctly predicted. Unlike previous research on clause, a correct item is considered to be correct only if both the start and end of the item are correctly predicted in the new ECSP task.

\subsection{Experimental Settings}
We use the BERT-Chinese\footnote{Available at: \url{https://github.com/huggingface/transformers}} model as the default backbone network, which using 12 layers, 768 -dimensional embeddings, 12 heads per layer, resulting in a total 110M parameters. Each span gets a span length feature $\Phi$ which is a learned 25 -dimensional vector representing the number of tokens in that span and each pair also gets a localized context length feature $\Psi$ which is twice as much as $\Phi$. we randomly divide the data with the proportion of 9:1, with 9 folds as training data and remaining 1 fold as testing data. The following results are reported in terms of an average of 10-fold cross-validation. We use Adam optimizer with a linear warmup and linear decay learning rate schedule and a peak learning rate of 5e-5. Dropout is applied with dropout rate 0.1 to all hidden layers of BERT and Classifiers. Mini-batch Size is 1 and early stopping of 20 evaluations on the dev set is used.

\subsection{Evaluation on the New ECSP Task}
\subsubsection{Overall Performance}
Table \ref{result:ecsp} shows our proposed ETC model performances with different span lengths on four sub-tasks: (emotion expression span extraction (EESE), emotion cause span extraction (ECSE), emotion-cause span-pair extraction (ECSPE), and emotion-cause span-pair extraction and classification (ECSP)).

Given a document with a $T$ token, there may be $N=T(T+1)/2$ spans. The huge search space makes the task extremely challenging. In this experiment, we created a length-restricted span (rather than just token) representation that achieves a dual goal: to improve memory efficiency and capture the majority (more than 98\% of emotions, see Table \ref{corpus}) for the span considered.

Compared with ETC-5 and ETC-15, ETC-20 gets great improvements on the ECSP task as well as the two sub-tasks. Specifically, we find that the improvements are mainly in the recall rate on the ECSE task, which finally lead to the great improvement in the recall rate of ECSP. The performance of the model does not decrease sharply as the length of the annotation increases, and our chosen span search scheme is far more memory efficient than a naive search over all possible spans in the input document. Yet our scheme still considers more than 98\% of all annotation. Our scheme is linear in the document length, not quadratic; because we limit our proposed ETC model to spans that are wholly in a document and have a max length of $L$ = 20 tokens.

In addition, the model achieved excellent F1 score 88.71 on ESE, but the F1 score on ECSP is 3.14\% lower than ECSPE, which indicates that it is not enough to extract the emotion without identifying the emotion categories. The presence of emotion clause does not necessarily convey emotional information explicitly, and emotions need to be classified.

\begin{table}[!t]
\centering
\small 
\begin{tabular}{lcccccccccccc}
\hline
\multicolumn{1}{c}{\multirow{2}{*}{Model}} & \multicolumn{3}{c}{EESE} & \multicolumn{3}{c}{ECSE} & \multicolumn{3}{c}{ECSPE} & \multicolumn{3}{c}{ECSP} \\ 
\multicolumn{1}{c}{} & P & R & F1 & P & R & F1 & P & R & F1 & P & R & F1 \\ \hline
ETC-5  &  87.78&  89.69&  \textbf{88.71}&  63.73&  10.48&  17.93&  56.19&  09.31&  15.90&  54.14&  08.90&  15.22\\
ETC-10 &  87.81&  89.53&  88.65&  61.05&  40.69&  48.81&  54.18&  37.23&  44.10&  50.78&  35.14&  41.50\\
ETC-15 &  86.78&  90.10&  88.39&  61.91&  51.78&  56.22&  53.66&  47.35&  50.20&  50.60&  44.40&  47.14\\
ETC-20 &  87.56&  89.31&  88.42&  60.23&  57.78&  \textbf{58.90}&  53.19&  51.11&  \textbf{52.11}&  49.91&  48.11&  \textbf{48.97}\\ \hline
\end{tabular}
\caption{Experimental results of all proposed ETC model and variants, where ETC-$n$ represents the maximum span length of the model is $n$.}
\label{result:ecsp}
\end{table}

\subsubsection{Effect of Localized Context}
As is shown in Table \ref{result:LC}, localized context can effective slightly improve the performance of the model. The localized context takes advantage of all information between two span, so it is able to enrich the source information when the model predicts the emotion labels, which leads to the performance of the model effective significantly improved.

\begin{table}[!t]
\centering
\small 
\begin{tabular}{lcccccccccccc}
\hline
\multicolumn{1}{c}{\multirow{2}{*}{Model}} & \multicolumn{3}{c}{EESE} & \multicolumn{3}{c}{ECSE} & \multicolumn{3}{c}{ECSPE} & \multicolumn{3}{c}{ECSP} \\ 
\multicolumn{1}{c}{} & P & R & F1 & P & R & F1 & P & R & F1 & P & R & F1 \\ \hline
Without &  87.08&  89.11&  88.07&  58.76&  57.10&  57.79&  52.73&  520.03&  51.31&  49.62&  47.24&  48.35\\
With    &  87.56&  89.31&  \textbf{88.42}&  60.23&  57.78&  \textbf{58.90}&  53.19&  51.11&  \textbf{52.11}&  49.91&  48.11&  \textbf{48.97}\\
\hline
\end{tabular}
\caption{Experimental results with localized context effects. ``With'' represents the localized context, ``Without'' means no localized context.}
\label{result:LC}
\end{table}

\subsection{Evaluation on the Traditional Task}
By relaxing the ECSP task to the clause-level, we further examine our model by comparing it with state-of-the-art of the traditional ECE and ECPE task.
\subsubsection{Baselines}
We employ a hierarchical Bi-LSTM network \textbf{Indep} proposed by \newcite{xia2019emotion} as baseline in ECPE task. The lower layer consists of a set of word-level Bi-LSTM modules, each of which corresponds to one clause, and accumulate the context information for each word of the clause. Attention mechanism is then adopt to get a clause representation. The upper layer consists of two components: one for emotion expression extraction and another for cause extraction. Each component is a clause-level Bi-LSTM which receives the independent clause representations and finally feed to the softmax layer for emotion prediction and cause predication. It has two interactive variants: \textbf{Inter-CE}, where the predictions of cause extraction are used to improve emotion extraction, and \textbf{Inter-EC}, where the predictions of emotion extraction are used to enhance cause extraction.

In addition to baselines mentioned above, we also considered several state-of-the-art methods and models in ECE task that need to provide annotations of emotional expressions in the test set in advance to evaluate the results of our proposed ETC model: \textbf{RB} is a rule based method \cite{lee2010text}; \textbf{CB} is common-sense based method \cite{russo2011emocause}; \textbf{ConvMS-Memnet} considers emotion cause analysis as a reading comprehension task and designs a multiple-slot deep memory network to model context information \cite{gui2017question}. \textbf{CANN} uses a co-attention neural network to identify emotion causes \cite{li2018co} and \textbf{CANN-E} eliminates the dependence of CANN on emotion annotation in the test data. \textbf{HCS} is proposed by \newcite{yu2019multiple} using a multiple-level hierarchical network to detect the emotion causes. \textbf{MANN} is the current state-of-the-art method employing a multi-attention-based model for emotion cause extraction \cite{li2019context}.

%%%%%%%%%%%%%%%%%%%%%%%%%%%%%%%%%%%%%%%%%%%%%%%%%%%%%%%%%%%
\subsubsection{Results and Analysis}
\begin{table}[!tbp]
    \begin{minipage}{0.5\textwidth}
        \subfloat[Performance on the EEE.]
        {
            \begin{tabular}{lccc}
            \hline
            Model&  Precision&  Recall&  F1\\\hline
            Indep&  83.75&  80.71&  82.10\\
            Inter-CE&  84.94&  51.22&  83.00\\
            Inter-EC&  83.64&  81.07&  82.30\\\hline
            ETC-Clause&  94.04&  95.13&  \textbf{94.57}\\
            \hline
            \end{tabular}
        }\\
        \subfloat[Performance on the ECPE.]
        {
            \begin{tabular}{lccc}
            \hline
             Model&  Precision&  Recall&  F1\\\hline
            Indep&  68.32&  50.82&  58.18\\
            Inter-CE&  69.02&  51.35&  59.01\\
            Inter-EC&  67.21&  57.05&  61.28\\\hline
            ETC-Clause&  88.17&  84.07&  \textbf{86.05}\\
            \hline
            \end{tabular}
        }
    \end{minipage}
    \begin{minipage}{0.5\textwidth}
        \subfloat[Performance on the ECE.]
        {
            \begin{tabular}{lccc}
            \hline
            Model&  Precision&  Recall&  F1\\\hline
            RB&  67.47&  42.87&  52.43\\
            CB&  26.72&  71.30&  38.87\\
            ConvMS-Mement&  70.76&  68.38&  69.55\\
            CANN&  77.21&  68.91&  72.66\\
            HCS&  73.88&  71.54&  72.69\\
            MANN&  78.43&  75.87&  77.06\\\hline
            MANN-E&  48.26&  31.60&  37.97\\
            Indep&  69.02&  56.73&  62.05\\
            Inter-CE&  68.09&  56.34&  61.51\\
            Inter-EC&  70.41&  60.83&  65.07\\\hline
            ETC-Clause&  91.29&  87.98&  \textbf{89.57}\\
            \hline
            \end{tabular}
        }
    \end{minipage}
    \caption{Comparison between our proposed ETC model and baselines on clause-level.}
    \label{result:clause}
\end{table}

The past clause-level models regarded the ECE task as a set of independent clause classification problems. By observing the Table \ref{result:clause} (c), we found that the proportions of emotion cause clauses and non-emotion-cause clauses were 18.36\% and 81.64\%, respectively. It is a serious class-imbalance classification problem and the model tends to predict the clause as non-emotion-cause more often. This is also the reason why their Recall scores were quite low (the highest was 75.87).

By contrast, it can found in Table \ref{result:clause} (c) that our proposed ETC model is absolutely higher on each indicator than the other baselines and no need to manually annotate the test set. This is because they can capture the relations of multiple clauses which help inferring the current clause. For example, if no other clauses in a document have been detected as an emotion cause, the model will increase the probability of the current clause being predicted as an emotion cause. This finally increases the Recall score. It is clear that by removing the emotion annotations (CANN-E), the F1 score of CANN drops dramatically (about 34.69\%). In contrast, our method does not need the emotion annotations and achieve 89.57\% in F1 score, which significantly outperforms the CANN-E model by 51.6\%.

\newcite{xia2019emotion} guessed that the expression clause extraction and cause clause extraction are not mutually independent. On the one hand, providing emotions can help better discover the causes; on the other hand, knowing causes may also help more accurately extract emotions. Our proposed ETC model uses a classifier to complete the classification of expression and cause, forcing the classifier to learn the intrinsic relationship between them. Thanks to BERT's self-attention mechanism, our proposed ETC model can capture the relationship between multiple clauses. It can found in Table \ref{result:clause} (b) that our proposed ETC model has been greatly improved on both expression clause extraction and cause clause extraction tasks. Compared with Indep, Inter-CE and Inter-EC, our proposed ETC model gets great improvements on the ECPE task as well as the two sub-tasks. Our span-based model achieves 11.57\%, 24.77\% and 24.5\% absolute gains on three sub-task compared to the best classification model, indicating the efficacy of our proposed ETC model.

\section{Related Work}
First of all, our work is related to extracting causes based on emotions expression presented in documents, i.e., emotion cause extraction (ECE). ECE was first proposed by \newcite{lee2010text}, given the fact that an emotion is often triggered by cause events and that cause events are integral parts of emotion, they proposed a linguistic-driven rule-based system for emotion cause detection. To solve the insufficient of no formal definition about event in emotion cause extraction and there was no open corpus available for emotion cause extraction, \newcite{gui2016event} released a corpus and re-formalized the ECE task as a clause classification problem. This corpus has received much attention in the following study and has become a benchmark corpus for ECE task research. Based on this corpus, several traditional rule-based models \cite{lee2010text,russo2011emocause,gui2014emotion}, machine learning models \cite{gui2016event,gui2017question,xu2017ensemble} and deep learning models \cite{gui2017question,li2018co,yu2019multiple,xia2019rthn,li2019context} were proposed. Recently, To solve the shortcoming of emotion expression must be annotated before cause extraction in the test set, \newcite{xia2019emotion} proposed emotion-cause pair extraction (ECPE) task, which aims to extract all potential clause-pairs of emotion expression and corresponding cause in a document.

\section{Conclusions and Future Work}
The key idea of task and model is to build span-based feature representation for emotion expression and causes to efficiently extract document information. Furthermore, our proposed ETC model is able to utilize the information based on an overall understanding of the document and a better localized context of interactions between spans. Comprehensive empirical studies demonstrate the effectiveness of our proposed ETC model. Since our proposed ETC model has a single input structure, so in the future we will explore how to incorporate discourse graphs into our proposed ETC model to further improve performance, and we intend to annotate a large-scale emotion-cause span-pair corpus to facilitate research.

\bibliographystyle{coling}
\bibliography{coling2020}

\end{document}